\documentclass[a4paper, 12pt]{article} 

\usepackage{times}  
\usepackage{url}  
\usepackage{graphicx} 
\usepackage{cite}
\usepackage{amsfonts}
\usepackage{fullpage}
\usepackage{authblk}

\usepackage{xspace}
\usepackage[dvipsnames, table]{xcolor}
\usepackage{amsmath}
\usepackage{rotating}
\usepackage{lipsum}
\usepackage[ruled,linesnumbered]{algorithm2e}
\usepackage{adjustbox}
\usepackage{multirow}
\usepackage{multicol}
\usepackage{hyperref}
\hypersetup{
  colorlinks   = true, %Colours links instead of ugly boxes
  urlcolor     = blue, %Colour for external hyperlinks
  linkcolor    = blue, %Colour of internal links
  citecolor   = red %Colour of citations
}

\usepackage[T1]{fontenc}
\usepackage[]{lmodern}
\usepackage[utf8]{inputenc}
\usepackage[english]{babel}

\usepackage[]{siunitx}
\def\C#1{\multicolumn{1}{c|}{#1}}

\def\V#1{\multicolumn{1}{c}{#1}}

\newcommand{\microrts}{\si{\micro}RTS\xspace}
\newcommand{\POAdaptive}{\textsc{POAdaptive}\xspace}
\newcommand{\ie}{\textit{i.e.}}
\newcommand{\csp}{\textsc{CSP}\xspace}
\newcommand{\cop}{\textsc{COP}\xspace}

\begin{document}

\title{Constrained optimization under uncertainty\\for decision-making problems:\\Application to Real-Time Strategy games}

\author[1]{Valentin Antuori}
\author[1,2]{Florian Richoux}
\affil[1]{LS2N, Université de Nantes, France}
\affil[ ]{\it \{first.last\}@univ-nantes.fr}
\affil[2]{JFLI, CNRS, National Institute of Informatics, Japan}
%\{first.last\}@univ-nantes.fr}
% \author{Valentin Antuori and Florian Richoux\\\textit{LS2N}, \textit{Université de Nantes}, France\\
% \{first.last\}@univ-nantes.fr}

\maketitle

\begin{abstract}
  Decision-making   problems   can   be   modeled   as   combinatorial
  optimization problems with Constraint Programming formalisms such as
  Constrained   Optimization   Problems.   However,   few   Constraint
  Programming  formalisms   can  deal   with  both   optimization  and
  uncertainty at  the same time,  and none  of them are  convenient to
  model problems  we tackle in this  paper. Here, we propose  a way to
  deal  with  combinatorial  optimization problems  under  uncertainty
  within the classical Constrained  Optimization Problems formalism by
  injecting the Rank Dependent Utility  from decision theory.  We also
  propose a proof of concept of our method to show it is implementable
  and  can solve  concrete  decision-making problems  using a  regular
  constraint  solver,  and  propose  a  bot  that  won  the  partially
  observable track of  the 2018 \microrts AI  competition.  Our result
  shows it is  possible to handle uncertainty  with regular Constraint
  Programming  solvers,  without  having  to define  a  new  formalism
  neither to  develop dedicated solvers.  This  brings new perspective
  to tackle uncertainty in Constraint Programming.
\end{abstract}

\section{Introduction}

Decision-making problems can be  modeled as combinatorial optimization
problems  through a  given  formalism,  and then  can  be solved  with
appropriated tools, \ie,  solvers. Combinatorial optimization problems
are  very frequent  problems in  domains such  as logistics,  finance,
supply  chain,   planning,  scheduling  and  in   industries  such  as
pharmaceutical industry, transportation,  manufacturing and automotive
industry~\cite{handbookCP}.

Strategy  games propose  a rich  environment to  study decision-making
problems, allowing  researchers to develop new  algorithmic approaches
to  model and  solve such  problems.   This is  particularly true  for
Real-Time Strategy games, or RTS games, offering a dynamic environment
under a fog  of war forbidding players to have  a complete information
about  the game  state.   Such environments  contain many  challenging
combinatorial optimization problems.

Combinatorial optimization problems can be expressed through different
formalisms.   One  convenient  formalism  used  in  AI  is  Constraint
Satisfaction  Problems (\csp)  and  Constrained Optimization  Problems
(\cop).  The  first formalism  deals with satisfaction  problems, \ie,
problems where all solutions have the  same quality.  In this paper, a
solution is  an assignment of each  variable of the problem  such that
all constraints are  satisfied.  The second formalism  \cop deals with
optimization problems, \ie, problems where there is a criteria to rank
solutions.

There  exist  many extensions  of  the  \csp formalisms  dealing  with
uncertainty,  but  very few  of  them  have  been extended  to  handle
optimization  problems,  and when  they  did,  they force  to  declare
additional parameters that might be undesirable and inconvenient while
modeling a problem.

This  paper  proposes   a  way  to  deal  with  a   specific  kind  of
decision-making  problems  through  combinatorial  optimization  under
uncertainty  within  the  classical  \cop  formalism  using  the  Rank
Dependent Utility from decision theory.  We exhibit a proof of concept
with  a simple  bot  playing to  the \microrts  game  while solving  a
decision-making problem of  choosing the right units  to produce.  Our
bot has  won the partially observable  track of the 2018  \microrts AI
competition.

This paper is organized as follows:  We first motivate why we focus on
single-stage decision-making  problem and  why   uncertainty   is
exclusively in the objective function in Section~\ref{sec:motivation}.   Then, we
introduce  basic notions  about  Constraint  Programming and  Decision
Theory         in         Section~\ref{sec:preliminaries}.          In
Section~\ref{sec:contrib}, we  expose our main contribution:  a way to
handle uncertainty within the classical  \cop formalism using the Rank
Dependent   Utility  and   finally  give   a  proof   of  concept   in
Section~\ref{sec:xp} by modeling with a \cop a decision-making problem
under  uncertainty  in  \microrts.   Related works  can  be  found  in
Section~\ref{sec:related}.           We           conclude          in
Section~\ref{sec:conclusion}.

\section{Motivation}\label{sec:motivation}

We introduce  in this section  the type of  problems we focus  on, and
motivate why such problems worth to be specifically tackled.

In this paper, we study  single-stage decision-making problem an agent
must solve under uncertainty, where  uncertainty lies on the value of
some stochastic variables  controlled by a third-party  agent (such as
the environment where our agent  evolves). Such stochastic values only
have an impact  on the objective function the agent  tries to maximize
or minimize, and not on constraints it must satisfy.

We  think  important  to  motivate   the  two  following  points:  why
single-stage decision-making  problems only, and why  only considering
uncertainty  on  the  objective  function  rather  than  on  both  the
objective function and the constraints.

Studying single-stage  decision-making problems means that  a decision
must be made  before revealing stochastic values so  far unknown. Once
these values are known, the agent can only observe the consequences of
its decision without having the possibility  to sharpen or fix it like
in  multi-stage   decision-making  processes.    Although  multi-stage
decision-making problems  are interesting  and would deserve  a proper
study,  we  think  single-stage  decision-making  problems  are  still
relevant and  capture all one-shot decision-making  problems that must
be made  recurrently. Concrete  examples can be  1. a  factory manager
deciding about  the production  of the month  taking into  account the
stock (known) and client orders (unknown), 2. blind auctions where one
aims  to win  some auctions  taking into  account the  available money
(known)  and  other participants  bid  (unknown)  or 3.   air  traffic
management  where one  must take  into account  the number  of waiting
planes  for  taking  off  and   landing  (known)  and  future  demands
(unknown).  In his PhD  thesis~\cite{Piette16}, Éric Piette shows that
decision-making problems in  strategy games can be  handle in practice
by single-stage decision-making problems only. Finally, another reason
to   study  single-stage   decision-making  problems   is  that   some
environments  do not  allow multi-stage problems:  To  do multi-stage
decision-making, some  stochastic variables  must be revealed  at each
stage. However, it  is easy to find natural  problems where stochastic
variables are never completely revealed. This is the case in RTS games
for instance, where the fog of war is never completely dissipated. The
problem we tackle in the paper belongs to this category.

Considering uncertainty  having only an  impact on a  solution quality
(the objective function) rather than its possibilities (the constraints)
makes sense  for the same  reasons as  above: There are  many concrete
decision-making problems where one knows  what is possible and what is
not, but does not  known what the quality of its  decisions will be. In
other  words,  the scope  of  our  possible  decisions is  known  (our
constraints) but  we live in  an uncertain, dynamic  environment where
events out  of our  control can  impact not  the applicability  of our
decisions but their quality. Examples  cited in the previous paragraph
are still  relevant here: whatever  our client  orders, we can  make a
production plan  of the month  regarding our available stock  only; we
can bid to an auction regarding only our available money, but we can
lose because of  better bids; and we can plan  air traffic knowing the
current situation,  but we can be  overwhelmed by a group  of arriving
planes if we made bad runway assignments.

\section{Preliminaries}\label{sec:preliminaries}

\subsection{Constraint Programming}\label{sec:cp}

The  basic  idea  behind  Constraint   Programming  is  to  deal  with
combinatorial problems by  splitting them up into  two distinct parts:
the first part is modeling your problem via one Constraint Programming
formalism.  This is  usually done by a human being  and this task must
be ideally  easy and intuitive.   The second part consists  in finding
one  or several  solutions based  on  your model.  This is  done by  a
solver, \ie, a program running without any human interventions.  % All
% the intelligence must be placed into this second part, and this is the
% main  reason   why  Constraint  Programming  is   part  of  Artificial
% Intelligence.

The  two  main  formalisms  in Constraint  Programing  are  Constraint
Satisfaction  Problems (\csp)  and  Constrained Optimization  Problems
(\cop). The difference between a \csp and a \cop is simple:

A  \csp  models a  satisfaction  problem,  \ie,  a problem  where  all
solutions are equivalent;  the goal is then to just  find one of them,
if any.   For instance:  finding a  solution of  a Sudoku  grid.  Good
grids lead to a unique  solution, but let's consider several solutions
are  possible  for  a  given  grid.  Then,  finding  one  solution  is
sufficient, and no solutions seem  better than another one. Sometimes,
we  may also  be  interested in  finding all  solutions  of a  problem
instance.

A \cop models an optimization problem, where some solutions are better
than  others.  For  instance: Several  paths  may exist  from home  to
workplace, but one of them is the shortest.

Formally, a \csp is defined by a tuple ($V$, $D$, $C$) such that:
\begin{itemize}
\item $V$ is a set of variables,
\item $D$ is a domain, \ie, a set of values for variables in $V$,
\item $C$ is a set of constraints.
\end{itemize}

A constraint over  $k$ variables can be seen as  a function from $D^k$
to  $\{true, false\}$  to make  explicit what  combinations of  values
among its $k$ variables are allowed or not.

Notice that  $D$ should  formally be  the set of  the domain  for each
variable in $V$, thus  a set of sets of values.  However, it is common
to define the  same set of values  for all variables of  $V$, thus one
can simplify  $D$ to  be the set  of values each  variable in  $V$ can
take.

A \csp models a problem, and a problem instance is expressed by a \csp
formula, \ie, a  set of constraints applied on variables  in $V$ where
all constraints are linked by a logical \textit{and}. The goal is then
to attribute  a value in  $D$ for each variable  in $V$ such  that all
constraints in $C$ are satisfied, \ie, outputs \textit{true}.

A \cop is defined  by a tuple ($V$, $D$, $C$, $f$)  where $V$, $D$ and
$C$ represent the  same sets as a  \csp, and $f$ is  an {\it objective
  function} applied on  variables in $V$. The goal is  first to find a
solution, \ie, a value of each  variable such that all constraints are
satisfied, like for \csp, but moreover to find the solution minimizing
or maximizing the objective function $f$ among all possible solutions.

\csp and  \cop deal  with certain information  only. There  exist many
extensions  of the  \csp  formalisms dealing  with uncertainty:  Mixed
\csp, Probabilistic \csp,  Stochastic \csp, etc. We  invite the reader
to   look   at   surveys~\cite{VJ05}  and~\cite{Hnich2011}   on   this
topic. However, few are convenient  to model a decision-making problem
where  one does  know  what  his or  her  possible  choices are  (\ie,
variables,  domains  and  constraints  are known  and  fixed),  but  a
third-party agent (a person, an environment, etc) fixes
the values of some specific variables. These values are unknown at the
moment we  must make  a decision  and impact the  value output  by the
objective  function.  Stochastic  \csp~\cite{SCSP} is  the  most  well
adapted formalism to  model such problems, but with  the huge drawback
that  constraints  are  considered   to  be  chance-constraints,  \ie,
constraints  are  considered true  if  their  probability to  be  true
reaches a given threshold.  The main  problem with such a formalism is
that this threshold  must be provided by the human  being modeling the
problem,  and it  is  often unclear  in  practice how  to  fix a  good
threshold  value  for  a  given  problem. This  does  not  follow  the
Constraint Programming philosophy where problem models must be easy to
produce by a human being, without any arbitrary choices.

Moreover, while \cop are a trivial extension of \csp with an objective
function,  it  is  absolutely  not  clear  how  to  extend  constraint
satisfaction formalisms  under uncertainty  to deal  with optimization
problems. Indeed, to each solution of a problem can correspond several
possible objective  function values, due to  uncertainty on stochastic
variables,  and such  values depend on  the state  of an  environment
determining stochastic  variable values.  How is  it then  possible to
discriminate solutions between them?

To the best of our  knowledge,  no Constraint  Programming formalisms  without
chance-constraints   able  to   handle  optimization   problems  under
uncertainty    have   ever    been    proposed.     We   propose    in
Section~\ref{sec:contrib} a  way to  deal with uncertainty  within the
classical \cop  formalism, allowing  us to  solve such  problem models
with classical solvers.

\subsection{Decision Theory}\label{sec:rdu}

We consider the set $\mathcal{D}$ of  decisions an agent can take. The
goal is to define a  preference relation $\succeq_\mathcal{D}$ on this
set.  Preferring the  decision $d_1$ over $d_2$ means  to prefer $d_1$
consequences  over $d_2$  ones,  thus  we can  also  consider a  space
$\mathcal{X}$  of  consequences,  and   study  a  preference  relation
$\succeq_\mathcal{X}$  on  this space  in  such  a  way that  we  have
$d_1 \succeq_\mathcal{D}  d_2 \iff x_1 \succeq_\mathcal{X}  x_2$ where
$x_i$ is  the consequence of  the decision  $d_i$. However, we  do not
have  this  equivalence  anymore  when uncertainty  comes  into  play,
because we  are not  sure anymore  the decision $d$  will lead  to the
consequence $x$.

In uncertain environments, we consider  the set $S$ of possible states
of the  environment.  We  consider consequences to  be sets  of states
after making a decision. Thus, we have $\mathcal{X} = \mathcal{P}(S)$.

Utility-based theories  consider $P$  a probability  distribution over
$\mathcal{X}$, \ie,  a probability distribution over  sets of possible
states  in  $S$.   Let  $p_d$  be the  probability  following  $P$  of
obtaining  the  consequence $x_d  \in  \mathcal{X}$  after making  the
decision $d$.

We can then introduce the notion of  lottery. A lottery $l$ is a tuple
$(x_1,p_1;...;x_n,p_n)$  where $x_i$  is a  consequence and  $p_i$ its
associated probability, such  that $\sum_{i=1}^n p_i =  1$.  A lottery
is thus a sum-up of a decision  in the sense it represents the list of
possible   consequences   of   a  decision   with   their   associated
probabilities. Let $\mathcal{L}$ be the set of lotteries.  We can then
define    a     preference    relation     $\succeq_\mathcal{L}$    on
$\mathcal{L}$. How we define  $\succeq_\mathcal{L}$ exactly depends on
the decision  theory, but the  idea is  to bring back  the equivalence
$d_1 \succeq_\mathcal{D} d_2 \iff l_1 \succeq_\mathcal{L} l_2$.

There  exist different  works  on decision  theory  to establish  this
equivalence.  We have thus the notion of Expected Utility (EU) defined
by~\cite{von1944theory} in the game  theory framework.  However EU has
a limited power of expression since one can quickly derivate paradoxes
such as the  Allais Paradox violating the  independence axiom, telling
that if someone  has no preference between decisions A  and B, then he
or she must  still not have no the  preference if we mix A  and B with
some decision C.

% Wald's maximin  model~\cite{wald1949} escapes from the  Allais Paradox
% but fits better to unrepeated one-shot decisions, \ie, not repeated
% decisions where results  are accumulated.  The basic idea is  to aim a
% situation  maximizing  the minimal  gain.   Wald's  maximin model  has
% however a weak discriminant power.

Choquet Expected Utility  is a decision theory based  on capacities, a
notion generalizing probabilities.  A special case of Choquet Expected
Utility  restricted to  probability deformation  function is  the Rank
Dependent             Utility             (RDU)             introduced
by~\cite{RDU_quiggin_82,quiggin1993generalized}. RDU has more power of
expression than EU since it can explain the Allais Paradox. Unlike EU,
RDU  allow  to  model  attraction  or repulsion  to  risks  through  a
probability deformation  function. This can help  to modify on-the-fly
the   behavior  of   an  agent   taking  a   decision  regarding   its
environment. 

The   Rank   Dependent   Utility   is    then   a   way   to   compute
$\succeq_\mathcal{L}$, and then to evaluate and compare lotteries such
that $l_1  \succeq_\mathcal{L} l_2  \iff RDU(l_1) \geq  RDU(l_2)$. RDU
applied   to   the   lottery   $l$  is   the   function   defined   by
Equation~\ref{eq:rdu}.

\begin{figure*}[ht]
  \footnotesize
  \begin{equation}\label{eq:rdu}
    RDU(l)  =  u(x_1)   +  \big(u(x_2)  -  u(x_1)\big)*\phi\left(\sum_{i=2}^np_i\right)   +  \big(u(x_3)  -
    u(x_2)\big)*\phi\left(\sum_{i=3}^np_i\right) + \ldots + \big(u(x_n)  -  u(x_{n-1})\big)*\phi(p_n)
  \end{equation}
\end{figure*}

% \begin{equation*}
%   RDU(l) = u(x_1) + \sum_{i=2}^n \bigg[\phi\big( \sum_{k=i}^n p(x_k)\big) [u(x_i) - u(x_{i-1})] \bigg]
% \end{equation*}

\noindent
In  Equation~\ref{eq:rdu},  $u(x)$  is  a utility  function  over  the
consequence  space, intuitively  giving a  score to  consequences, and
$\phi(p)$  an  increasing function  from  $[0,  1]$  to $[0,  1]$  and
interpreted  as  a  probability  deformation  function.  The  function
$\phi(p)$  can  be anything,  as  soon  as  it  is monotone  and  both
equalities  $\phi(0)=0$  and  $\phi(1)=1$ hold.  Consequences  in  the
lottery $l$ are ordered such that $\forall  x_i, x_j$ with $i < j$, we
have $u(x_i) \leq u(x_j)$.

This  probability   deformation  function   $\phi$  allows   to  model
risk-aversion since a concave $\phi$ function defines an attraction to
risks and a convex $\phi$ function a repulsion to risks.  Intuitively,
if we  have $\phi(p)  \leq p$  for all  $p$, then  the agent  taking a
decision will underestimate  gains probabilities and then  will show a
kind of pessimism about risks.  We  will have the opposite behavior if
we have $\phi(p)  \geq p$ for all $p$. Notice  that sigmoid functions,
which  are neither  concave  nor  convex, are  also  possible. In  our
experiments in Section~\ref{sec:xp}, we  use a sigmoid function rather
than a convex  function to model pessimism,  to decrease probabilities
of good outcomes and increase probabilities of unfavorable ones.

Remember that consequences  $x_i$ in $l$ are ordered  according to the
value of $u(x_i)$, such that  consequences with a small score outputed
by the utility function $u$ are placed at the beginning of the lottery
$l$. The intuition behind Equation~\ref{eq:rdu} is then the following:
With probability $p=1$, by making  the decision $d$, you are sure to
have at least the score of the worst consequence $x_1$, \ie, $u(x_1)$.
Then, with (deformed) probability $\phi(p_2+\ldots+p_n)$, you can have
the score $u(x_1)$ plus a  gain equals to $\big(u(x_2) - u(x_1)\big)$.
With probability  $\phi(p_3+\ldots+p_n)$, you  can have  an additional
gain equals to $\big(u(x_3) - u(x_2)\big)$,  and so on until having an
additional  gain  equals  to   $\big(u(x_n)  -  u(x_{n-1})\big)$  with
probability $\phi(p_n)$. The  obtained value depends on  the order, or
rank,  of  the  value  of   the  utility  function applied  to
consequences, justifying the name ``Rank Dependent Utility''.

However, defining a utility function $u$ over the consequence space it
not  easy,  even  for  numerical-only  consequences.   This  space  is
completely dependent on  the problem and even on  the problem instance
so it  is not realistic  to propose general-purpose  utility functions
that could work and certify a  behavior on any kind of decision-making
problem.  This  is however  possible with the  probability deformation
function  $\phi$ since  it  is  always a  function  from  $[0, 1]$  to
$[0, 1]$.

Our decision-making problems being modeled as optimization problems, a
consequence  $x$ of  a  decision $d$  is the  value  of our  objective
function.  Therefore, the relation $\succeq_\mathcal{X}$ is merely the
relation $\geq$ over real numbers. This implies that $u$ is a function
from $\mathbb{R}$ to $\mathbb{R}$. In this work, we consider $u$ to be
the identity function $id(x) = x$ % (this corresponds to Yaari's dual
% theory~\cite{yaari_dual})
and will use generic probability deformation
functions $\phi$ to change an agent's behavior regarding risks.
% \begin{figure*}[ht]
%   \begin{equation}\label{eq:rdu}
%     RDU(l)  =  x_1   +  (x_2  -  x_1)*\phi(\sum_{i=2}^np_i)   +  (x_3  -
%     x_2)*\phi(\sum_{i=3}^np_i) + \ldots + (x_n  -  x_{n-1})*\phi(p_n)
%   \end{equation}
% \end{figure*}

\section{Main contribution}\label{sec:contrib}

The  main difficulty  to tackle  a combinatorial  optimization problem
under uncertainty via  Constraint Programming is the  lack of reliable
criterion to attribute a quality to each possible solution. How do you
rank solutions  if they  lead to  different objective  function values
regarding possible values of stochastic variables?

% As discussed in
% Section~\ref{sec:cp},   without   chance-constraints,  we   face   a
% bi-objective problem with an objective  function to optimize and the
% probability of satisfaction of constraints to maximize.

The main  contribution of this paper  is proposing to inject  the Rank
Dependent  Utility  from  decision  theory  into  the  classical  \cop
formalism  to   solve  optimization  problems  under   uncertainty.

We consider decision-making problem where one knows what our variables
are,  what values  they can  take  (\ie, we  know the  domain of  each
variable), what values combinations are  possible or not (\ie, we know
our constraints), but where we  have an objective function to optimize
implicating stochastic variables  for which values are  unknown at the
moment we  must take  a decision,  such that  only a  third-party (the
environment, an independent agent, etc) has the power to set the value
of these variables.

This describes in fact most common decision-making situations: when we
have to take a decision, we  often miss some pieces of information (we
cannot have a  perfect knowledge about everything) that  still have an
impact on  the quality of  our decision. Should  I invest my  money in
stocks or bitcoins?  We do not know if the price will climb up or fall
down, but we know however what we can or cannot do (how much money can
we invest  for instance).  The quality  of our  decision will  be only
revealed once stochastic variables values will be known.

\subsection{Injecting RDU into \cop}

We recall we are interested in modeling uncertainty in decision-making
problems. These  problems without  uncertainty can be  modeled through
the  \cop formalism.  For many  cases, uncertainty  in decision-making
problems does not affect what you can or cannot do but on external
unknown elements that have a direct impact on the decision quality.

An easy  way to  model such  decision-making problems  is to  model it
through the regular  \cop formalism, by defining 1. a  set of decision
variables, \ie, regular variables which the solver has the control on,
2. a set  of stochastic variables, representing all  unknown pieces of
information, 3. a  domain for both decision  and stochastic variables,
4.   a probability  distribution  for the  domain  of each  stochastic
variable, 5.  a  set of constraints upon decision variables  and 6. an
objective  function  mixing  decision and  stochastic  variables.   If
probability distributions of stochastic  variables are unknown, we can
approximate them  with statistics.  A convenient  point with  games is
that we  can often simulate  their environment or analyze  replays and
then collect those statistics fairly easily.

Like Equation~\ref{eq:rdu} suggests, we  need to know all consequences
of a decision  to compute RDU. This is of  course intractable since we
have $|D|^{|S|}$ consequences for each decision, with $|S|$ the number
of stochastic variables  and $|D|$ the cardinality of  their domain. A
convenient way  to approximate RDU  is to  do Monte Carlo  sampling of
stochastic    variable    values,    following    their    probability
distribution.

We can now apply  our objective function to compute the  RDU and get a
usable metric under uncertainty, which allows us to rank solutions and
guide our decisions.

Let's consider  a problem  modeled by a  \cop with  decision variables
$v_i$, stochastic variables $s_j$ and  an objective function $f$. Like
described  in   Subsection~\ref{sec:rdu},  consequences  $x_i$   of  a
decision $d$ corresponds to values output by $f$.  In the context of a
\cop, a decision  $d$ is a vector of values  assigned to each decision
variables $v_i$.   Using Algorithm~\ref{algo:rdu}, we can  compute the
relation  $\succeq_\mathcal{D}$ among  decisions by  approximating the
RDU  of  their  respective  lottery with  Monte  Carlo  samplings.

\begin{algorithm*}[ht]
  \SetKwData{Left}{left}\SetKwData{This}{this}\SetKwData{Up}{up}
  \SetKwFunction{Union}{Union}\SetKwFunction{FindCompress}{FindCompress}
  \SetKwInOut{Input}{input}\SetKwInOut{Output}{output}

  \Input{A decision $d$,  \ie, a vector in $D^n$, with  $D$ the domain
    of decision variables $v_1, \ldots, v_n$}
  \Output{A preference on $d$, \ie, a real value}
  \BlankLine
  Initialize an empty vector $x$ of size $k$, with $k$ a parameter for
  the number of wanted samples\;
  \For{$i=1$ \KwTo $k$}{
    Sample values for stochastic variables $s_1, \ldots, s_m$ according to their probability
    distribution\;
    \tcp{$f$  is  our objective  function,  taking  both decision  and
      stochastic variables}
    $x[i] \leftarrow f(v_1, \ldots, v_n, s_1, \ldots, s_m)$\;
  }
  Sort(x)\;
  \tcp{Considering each  sample has  a probability  $\frac{1}{k}$, computes
    RDU}
  RDU $\leftarrow x[1]   +  (x[2]  -  x[1])*\phi(\frac{k-1}{k})   +  (x[3]  -
  x[2])*\phi(\frac{k-2}{k}) + \ldots + (x[k]  -  x[k-1])*\phi(\frac{1}{k})$\;
  \Return{RDU}
  \BlankLine
  \caption{Estimating a preference on the decision $d$}
  \label{algo:rdu}
\end{algorithm*}

We give details about Algorithm~\ref{algo:rdu} here. It takes as input
a solution  $d$ (or a  decision), \ie, a vector  of values in  $D$ for
each decision variables of the  problem.  The algorithm outputs a real
number, a preference  on the decision, \ie, its  estimated RDU, giving
us  the  opportunity to  compare  it  with  other decisions.   Line  1
initializes a vector $x$ to save  $k$ values of the objective function
$f$,  each value  computed  with a  different  sampling of  stochastic
variables. From Line 2 to Line 5, we sample stochastic variable values
following their  probability distribution, computes the  values of $f$
regarding $d$ and sampled values, and  store them into the vector $x$.
This  vector  is  sorted  in  Line   6.  Lines  7  and  8  compute  an
approximation  of the  RDU applying  Equation~\ref{eq:rdu} and  return
this value.

In  our  experiments  next  section,   we  draw  $k=50$  samples.   In
Algorithm~\ref{algo:rdu}, $k$ being  a parameter and not  an input, we
do not  take it into count  to compute the algorithm  complexity.  The
complexity  of   Algorithm~\ref{algo:rdu}  is  then   in  $\Theta(f)$,
depending  on  the complexity  of  the  objective function  $f$  only.
Sampling $m$  stochastic variables  is also outside  the scope  of the
complexity of Algorithm~\ref{algo:rdu}  since stochastic variables are
not among its inputs.

\section{Proof of concept}\label{sec:xp}

We  give  a  proof of  concept  of  our  contribution  to show  it  is
implementable  and use  it to  solve a  decision-making problem  under
uncertainty  in a  RTS  game. We  have  included this  decision-making
solving system  into a bot  playing to  the game \microrts.   Our bot,
named \POAdaptive, has won the  partially observable track of the 2018
\microrts  AI competition  organized within  the CIG  2018 conference.
The  code of  our bot,  our  experimental setup  and our  experimental
results   can   be  found   in   the   following  github   repository:
\href{https://github.com/richoux/microrts-uncertainty/tree/v1.0}{github.com/richoux/microrts-uncertainty/tree/v1.0}.
%\textcolor{blue}{hidden for double-blind review process}. 

We  will consider  the following  problem: RTS  game propose  to train
units which often follow a  rock-paper-scissors scheme. Because of the
fog of war, we do not perfectly know the enemy army composition and we
must infer  his or  her strategy from  some partial  observations.  We
must constantly  take a  production decision answering  this question:
``What next units should I produce to counter my enemy strategy?''

\subsection{\microrts}

We decided to use \microrts has an experimental environment. \microrts
is  an open-source,  minimalist real-time  strategy game  developed by
Santiago Ontañón for research purpose~\cite{microRTS}.

\begin{figure}
  \centering
  \includegraphics[width=0.5\linewidth]{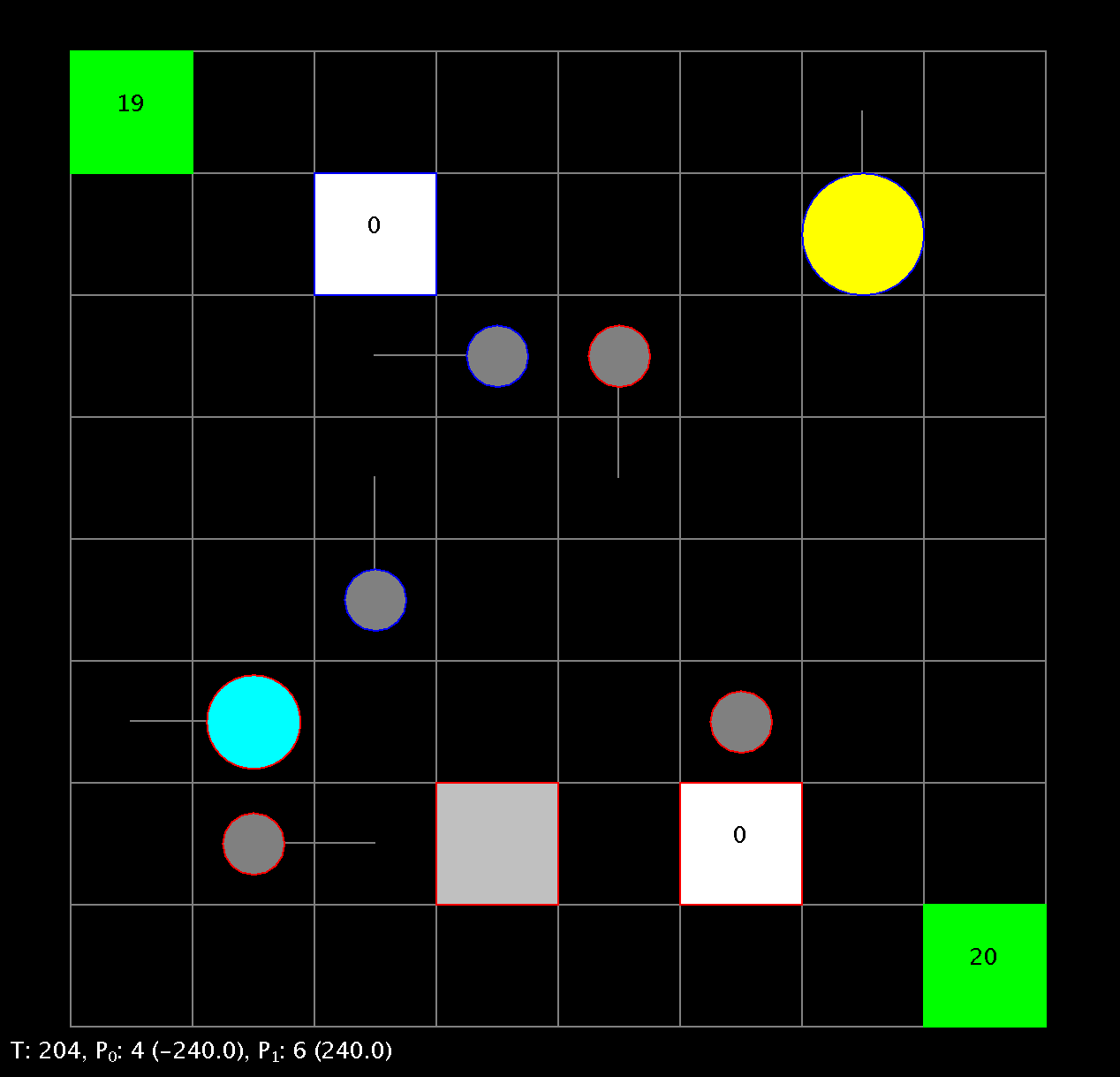}
  \caption{A game  frame of \microrts on  a 8x8 map. Resources  are in
    green, squares are buildings and round items are units.}
  \label{fig:microrts}
\end{figure}

The game  is made upon  classical RTS mechanisms: there  are resources
(or  money) to  gather (green  squares in  Figure~\ref{fig:microrts}).
This money allow us to build  buildings and train units. In \microrts,
there are two kind of buildings:  bases (white squares) where money is
stocked and  barracks (grey  squares) where army units  are produced.
Four units are  available in \microrts: workers  (small grey circles),
light     units     (orange      circles,     not     appearing     in
Figure~\ref{fig:microrts}),  ranged  units  (blue circles)  and  heavy
units (large yellow circles).  Workers  are weak against all units but
are  the only  ones  able  to gather  resources  and build  buildings.
Light,  ranged and  heavy  units are  following a  rock-paper-scissors
scheme, in the sense that heavy  units are strong against light units,
light units are strong against range  units and range units are strong
against heavy units.

To win, a player must destroy all enemy units and buildings. If nobody
reaches that goal before a fixed number  of frames, the game ends in a
draw.

\microrts supports  both complete  and partially observable  games. In
order  to  test  our  method  solving  decision-making  problem  under
uncertainty,  we used  \microrts exclusively  in partially  observable
mode. 

\subsection{Deciding about unit production}

We propose here a model of  our production problem through the regular
\cop  formalism. Let's  consider $\{H,  L, R\}$  the heavy,  light and
ranged type of units, respectively. We have:

\begin{itemize}
\item Two kind of decision variables: $plan_X$, with $X \in \{H, L,
  R\}$, representing the total number of  units of type $X$ we should have
  (\ie,  the total  number  of  units we  currently  possess plus  the
  number of units we plan to produce), and $assign_{XY}$, $\forall X,Y
  \in \{H,  L, R\}$, the number  of our units  of type $X$ we  plan to
  use to counter enemy units of type $Y$.
\item One kind of stochastic variables: $enemyUnits_X$, with $X \in \{H, L,
  R\}$, representing the  total number of units of type  $X$ the enemy
  currently possesses.
\item  Domains for  each  variable are  natural numbers  from  0 to  a
  threshold. We  used 20 as a  threshold in our experiments,  which is
  sufficient for small maps in \microrts.
\item Two kind of constraints:\\
  $assign_{HL} + assign_{HR} + assign_{HH} = plan_H$\\
  $assign_{RL} + assign_{RR} + assign_{RH} = plan_R$\\
  $assign_{LL} + assign_{LR} + assign_{LH} = plan_L$\\ \\
  $3(plan_H  -  ourUnits_H)  +  2(plan_R  -  ourUnits_R)  +  2(plan_L  -
  ourUnits_L) \leq stockResource$\\

  The  first three  constraints create  the bridge  between the  total
  number of units of  type $X$ we aim to have and  the number of units
  of type  $X$ we  consider we  need to counter  an unknown  number of
  enemy units of type $Y$. The last constraint is the resource balance
  constraint: given  $ourUnits_X$ the number  of units of type  $X$ we
  currently have, ($plan_X  - ourUnits_X$) corresponds to  the number of
  units $X$ we have to produce. A heavy unit costs 3 resource points, a
  light  and  ranged  units  only  2.   The  parameter  stockResources
  corresponds to the current resource  points we possess at the moment
  of we must decide about our production.
  
\item The objective function $\max target_H + target_L + target_R$ with
  \begin{equation*}
    \begin{split}
      target_X = \min\{1, ( HX * assign_{HX}\\+ RX * assign_{RX}\\ + LX *
      assign_{LX}\\- enemyUnits_X )\}
    \end{split}
  \end{equation*}
\end{itemize}

\noindent
where $X  \in \{H,  L, R\}$  and coefficient  of $AB$-type  (\ie, $HH$,
$HL$, $\ldots$,  $RR$) are  constants representing  how many  units of
type $A$ we need  to counter a unit of type $B$.  The min function for
$target_X$ is to avoid a mere sum of the expressions $HX * assign_{HX}
+ RX  * assign_{RX} + LX  * assign_{LX} - enemyUnits_X$  for the three
possible  $X \in  \{H,  L, R\}$,  otherwise it  would  lead to  simply
produce the unit  with the highest $AB$-type coefficient.  We take the
minimum between these expressions and the  value 1 to allow to produce
up to one more unit than necessary.

Our $AB$-type coefficients have been estimated by running 200 games of
10  units of  $A$ against  10 units  of $B$,  for each  combination of
$AB \in \{H, L,  R\}^2$.  We then took the ratio  of the total numbers
of  surviving units  over 200  simulations.  For  instance, after  200
games of  type ``10  heavy versus  10 light'',  we had  1284 surviving
heavy units and 480 surviving light units.  Then our parameter $HL$ is
equals to $\frac{480}{1284} = 0.3738$  (\ie, we need 0.3738 heavy unit
to   deal   against   1   light   unit)  and   $LH$   is   equals   to
$\frac{1284}{480} = 2.675$.

Finally, statistics  on $enemyUnits_X$ stochastic variables  have been
made  by   analyzing  800  replays   of  \microrts  games   from  2017
competitions.  For  each frame  and each unit  type, we  counted these
units occurrence.  These statistics are sharpen by observations while
playing a game: if we observe for  instance 3 enemy light units at the
same moment,  we nullify probabilities  that the enemy  has 0, 1  or 2
light units only, and we normalize remaining probabilities.

\subsection{Experiments}

Few \microrts bots have been developed for partially observable games,
and most  of them are  in fact scripted bots.   We have taken  a basic
rush  bot  and  only  modify   its  production  behavior,  giving  our
bot \POAdaptive. We did not modify its initial build order (produce no
additional  workers,  start immediately  a  barracks  with our  unique
worker and then  gather resources until the end of  the game). We only
add a  quick hit-and-run  behavior for  our ranged  units and  a light
seek-and-destroy behavior.

Our  bot adapts  its  production  in function  of  the RDU  preference
computed according  to the objective  function. We give our  pure \cop
model  to  the  GHOST   solver~\cite{RUB16},  a  solver  dealing  with
classical  \csp  or  \cop  models and  unable  to  handle  uncertainty
directly. This shows  that our way to inject decision  theory into the
classical \cop  formalism is sufficient  to handle uncertainty  and do
not required to develop a new formalism neither dedicated solvers.  We
give to the solver 100 milliseconds per frame as computation budget to
solve our \cop  problem, to be consistent  with \microrts competitions
rules.

\begin{figure}[ht]
  \centering
  \includegraphics[width=0.6\linewidth]{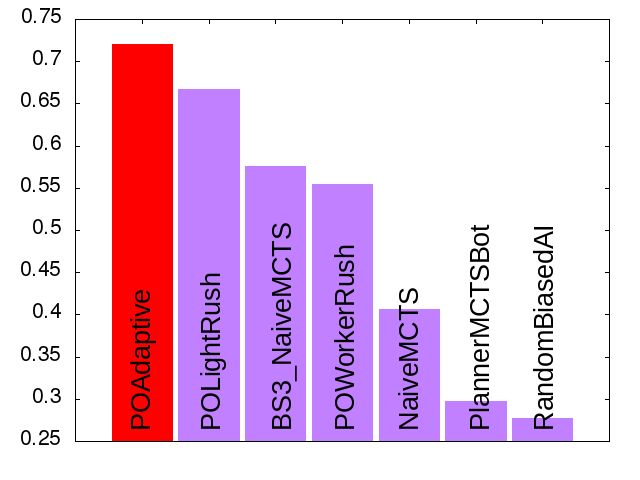}
  \caption{Final normalized  scores of the partially  observable track
    of the 2018 \microrts AI competition.}
  \label{fig:competition}
\end{figure}

With our bot \POAdaptive, we won the partially observable track of the
2018 \microrts  AI competition, over  7 competitors. We did  not tweak
our bot before the competition to be efficient on the competition maps
neither against specific  bots (such as 2017  competitors).  This show
that our decision-making solving is  efficient enough to beat scripted
rush bots  as well as  MCTS-based and Hierarchical  Task Network-based
bots.   Figure~\ref{fig:competition} shows  final scores  of the  2018
\microrts  AI  competition, over  720  games  for  each bot  within  a
round-robin tournament over 12 maps (4 of them were kept secret before
the beginning of the competition). The  cumulative score is the sum of
the score result for each game, \ie, 1  for a win, 0.5 for a tie and 0
for  a loss.  The scores  are  normalized by  dividing the  cumulative
scores by the number of games per bots, \ie, 720.

To evaluate our decision-making process, we run 100 games (50 starting
at the  North-East position, 50  starting at the  South-West position)
between the second  best bot of the competition,  POLightRush bot, and
four  methods:  \POAdaptive  using   RDU  with  a  pessimistic  $\phi$
function, \POAdaptive  using RDU  with an optimistic  $\phi$ function,
\POAdaptive using Expected Utility instead  of RDU (this can be easily
done by using RDU with $\phi$ as the identity function), and finally a
baseline bot  having exactly the  same behavior as  \POAdaptive except
for  the unit  production  decision, taken  randomly  among the  three
military  units.  The  pessimistic  function we  use  is the  logistic
function $\phi(p)  = \frac{1}{1 + exp(  - \lambda * (2*p  - shift) )}$
where  $p$ is  the probability  and with  parameters $\lambda=10$  and
$shift=1.3$.    The  optimistic   function  is   the  logit   function
$\phi(p)   =   1   +  \frac{log(   \frac{p}{2-p}   )}{\lambda}$   with
$\lambda=10$.

\begin{table}[ht]
  \centering{
    \begin{tabular}{|c|c|c|c|c|}
      \V{} & \V{} & \multicolumn{3}{c}{Map size}\\
      \cline{3-5}
      \V{} & \C{} & 8x8 & 12x12 & 16x16\\
      \hline
      \multirow{4}{*}{Baseline} & Win & 14 & 38 & 50\\
                   & \cellcolor{lightgray}Tie & \cellcolor{lightgray}0 & \cellcolor{lightgray}2 & \cellcolor{lightgray}12\\
                   & Loss & 86 & 60 & 38\\
                   & \cellcolor{lightgray}Score & \cellcolor{lightgray}14 & \cellcolor{lightgray}39 & \cellcolor{lightgray}56\\
      \hline
      \multirow{4}{*}{Expected utility} & Win & 27 & 35 & 52\\
                   & \cellcolor{lightgray}Tie & \cellcolor{lightgray}2 & \cellcolor{lightgray}6 & \cellcolor{lightgray}9\\
                   & Loss & 71 & 59 & 39\\
                   & \cellcolor{lightgray}Score & \cellcolor{lightgray}28 & \cellcolor{lightgray}38 & \cellcolor{lightgray}56.5\\
      \hline
      \multirow{4}{*}{RDU with optimistic $\phi$} & Win & 23 & 44 & {\bf 55}\\
                   & \cellcolor{lightgray}Tie & \cellcolor{lightgray}10 & \cellcolor{lightgray}7 & \cellcolor{lightgray}{\bf 13}\\
                   & Loss & 67 & 49 & {\bf 32}\\
                   & \cellcolor{lightgray}Score & \cellcolor{lightgray}28 & \cellcolor{lightgray}47.5 & \cellcolor{lightgray}{\bf 61.5}\\
      \hline
      \multirow{4}{*}{RDU with pessimistic $\phi$} &  Win & {\bf 26} & {\bf 50} & 57\\
                   & \cellcolor{lightgray}Tie & \cellcolor{lightgray}{\bf 6} & \cellcolor{lightgray}{\bf 5} & \cellcolor{lightgray}5\\
                   & Loss & {\bf 68} & {\bf 45} & 38\\
                   & \cellcolor{lightgray}Score & \cellcolor{lightgray}{\bf 29} & \cellcolor{lightgray}{\bf 52.5} & \cellcolor{lightgray}59.5\\
      \hline
    \end{tabular}
  }

  \vspace{0.25cm}  
  \caption{Results of 100 games played against LightRush bot on three small maps.  In
    bold, results with the highest score for each map.}
  \label{tab:xp}
\end{table}

\begin{figure}[ht]
  \centering
  \includegraphics[width=0.5\linewidth]{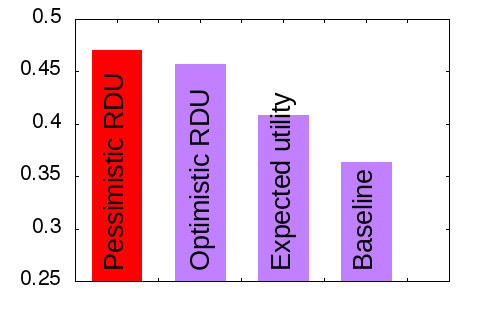}
  \caption{Normalized sum of scores of Table~\ref{tab:xp} (small maps) for each method.}
  \label{fig:xp}
\end{figure}

\begin{figure}[hb]
  \centering
  \includegraphics[width=0.5\linewidth]{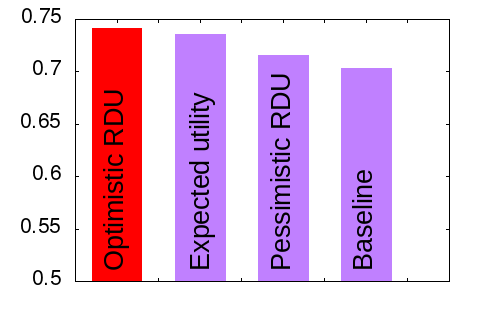}
  \caption{Normalized sum of scores of Table~\ref{tab:xp_large} (large maps) for each method.}
  \label{fig:xp_large}
\end{figure}

We run experiments  on three small basic maps 8x8,  12x12 and 16x16 as
well as on  three large basic maps 24x24, 32x32  and 64x64. Results on
small maps are shown in  Table~\ref{tab:xp}, and normalized scores are
illustrated   by   Figure~\ref{fig:xp}.    These  results   are   more
representative  of our  decision-making system  performances on  small
maps.  Indeed  on larger  maps, \POAdaptive has  too few  occasions to
meet enemy units since no scouting  behavior has been written (we know
by experience that  coding a proper scouting behavior would  not be so
trivial).   Thus, behavior  of  these  four methods  tends  to be  the
same. \POAdaptive's adaptation  skills give it a  slight advantage and
then slight  better scores than  the baseline method. On  these larger
maps,  RDU with  the optimistic  $\phi$ gives  the best  results (this
optimistic version  was besides  used for  the CIG  2018 competition),
slightly better than  the pessimistic version.  Results  on large maps
are shown in Table~\ref{tab:xp_large} and Figure~\ref{fig:xp_large}.

\begin{table}[ht]
  \centering{
    \begin{tabular}{|c|c|c|c|c|}
      \V{} & \V{} & \multicolumn{3}{c}{Map size}\\
      \cline{3-5}
      \V{} & \C{} & 24x24 & 32x32 & 64x64\\
      \hline
      \multirow{4}{*}{Baseline} & Win & 59 & 56 & 21\\
                   & \cellcolor{lightgray}Tie & \cellcolor{lightgray}34 & \cellcolor{lightgray}37 & \cellcolor{lightgray}79\\
                   & Loss & 7 & 7 & 0\\
                   & \cellcolor{lightgray}Score & \cellcolor{lightgray}76 & \cellcolor{lightgray}74.5 & \cellcolor{lightgray}60.5\\
      \hline
      \multirow{4}{*}{Expected utility} & Win & 60 & {\bf 62} & 28\\
                   & \cellcolor{lightgray}Tie & \cellcolor{lightgray}37 & \cellcolor{lightgray}{\bf 35} & \cellcolor{lightgray}70\\
                   & Loss & 3 & {\bf 3} & 2\\
                   & \cellcolor{lightgray}Score & \cellcolor{lightgray}78.5 & \cellcolor{lightgray}{\bf 79.5} & \cellcolor{lightgray}63\\
      \hline
      \multirow{4}{*}{RDU with optimistic $\phi$} & Win & {\bf 71} & 63 & 24\\
                   & \cellcolor{lightgray}Tie & \cellcolor{lightgray}{\bf 21} & \cellcolor{lightgray}32 & \cellcolor{lightgray}76\\
                   & Loss & {\bf 8} & 5 & 0\\
                   & \cellcolor{lightgray}Score & \cellcolor{lightgray}{\bf 81.5} & \cellcolor{lightgray}79 & \cellcolor{lightgray}62\\
      \hline
      \multirow{4}{*}{RDU with pessimistic $\phi$} &  Win & 66 & 54 & {\bf 27}\\
                   & \cellcolor{lightgray}Tie & \cellcolor{lightgray}25 & \cellcolor{lightgray}38 & \cellcolor{lightgray}{\bf 73}\\
                   & Loss & 9 & 8 & {\bf 0}\\
                   & \cellcolor{lightgray}Score & \cellcolor{lightgray}78.5 & \cellcolor{lightgray}73 & \cellcolor{lightgray}{\bf 63.5}\\
      \hline
    \end{tabular}
  }

  \vspace{0.25cm}  
  \caption{Results of 100 games played against LightRush bot on three large maps.  In
    bold, results with the highest score for each map.}
  \label{tab:xp_large}
\end{table}

On small  maps, Table~\ref{tab:xp}  and Figure~\ref{fig:xp}  show that
both RDU  versions outperformed  the EU version,  itself outperforming
our baseline.   Unlike for large  maps, the pessimistic  version gives
slightly  better results  than the  optimistic version.   This can  be
explained by the fact that small maps do not give you a lot of time to
react  when you  spot an  unfavorable enemy  army composition.   Being
already prepare to the worst helps in that case.

\section{Related works}\label{sec:related}

Uncertainty  has been  intensively studied  for the  last 30  years in
fields  dealing with  combinatorial optimization  such as  Operational
Research~\cite{Birge2011}, but significantly less works have been done
for     optimization      under     uncertainty      in     Constraint
Programming~\cite{Hnich2011}.   To the best of  our knowledge,  all methods  to
solve optimization problems  through Stochastic Constraint Programming
use  a formalism  considering chance-constraints,  usually handled  by
scenario-based methods.% , thus making harder and
% even  arbitrary  to model  a  problem  in  practice, as  discussed  in
% Section~\ref{sec:preliminaries}
\cite{Babaki2017}  is a  recent  example where  the  authors inject  a
probabilistic inference engine from the graphical model community into
a classical  solver to solve  Stochastic \csp instances,  thus dealing
with chance-constraints.  % They avoid the
% problem of uncertainty in the objective function value by reporting it
% into a  chance-constraint that  must be  satisfied with  a probability
% exceeding a given threshold.  Their paper proposes to use results from
% a research field outside the Constraint Programming community. In this
% paper, we do the same by  injecting results from decision theory, what
% has never been proposed to the best of our knowledge in Constraint Programming.

There are also  few works in RTS Game AI  using Constraint Programming
techniques, in particular through Constraint Satisfaction/Optimization
Problems, and few  of them dealing with uncertainty.   However, we can
cite   works   of~\cite{Koriche2016,KoricheLPS16,KoricheLPT17}   where
authors use Stochastic \csp to make a bot participating to the General
Game Playing competition. Their bot has won the 2016 competition.

Although the following  papers do not deal with  uncertainty, they all
focus on  solving optimization  problems in  RTS games,  in particular
StarCraft.  Thus \cite{RUO14}  propose to model with  \cop the optimal
building placement to make a wall at  a base entrance in order to make
easier its defense. \cite{FR15,RUB16} propose  a \csp and \cop solver,
GHOST, that we used for  our experiments. Their Constraint Programming
solver as  been designed to  output good quality solution  within some
tenth of milliseconds, make it usable in RTS games.

Beyond  Constraint Programming  but close  enough, \cite{ChurchillB11}
use a branch  and bound algorithms to optimize build  order in the RTS
game  StarCraft.   Like~\cite{RUO14},   \cite{Certicky13}  tackle  the
problem  to optimize  a wall-in  building placement  in StarCraft  but
through the prism of Answer-Set Programming.

\section{Conclusion}\label{sec:conclusion}

In  this  paper,  we  proposed   a  way  to  deal  with  combinatorial
optimization   problems  under   uncertainty   within  the   classical
Constrained  Optimization Problems  formalism  by  injecting the  Rank
Dependent Utility from decision theory.  The difficulty for Constraint
Programming formalisms  of handling both optimization  and uncertainty
at the  same time was  due to the  impossibility to rank  solutions if
they lead  to different  objective function values  regarding possible
values of stochastic variables.

% was  complex: either  you  adopt a  Stochastic Constraint  Programming
% using chance-constraints, such as Stochastic \csp, by you must declare
% a  threshold  parameter  in  you  model which  is  often  in  practice
% arbitrary fixed, or you face a bi-objective problem were you must both
% optimize  your  objective function  and  maximize  the probability  of
% satifaction of your constraints.

We get around this difficulty  by computing preferences over decisions
with the Rank Dependent Utility  using our objective function to score
decisions. This allow us to show  it is possible to handle uncertainty
with regular Constraint Programming  solvers, without having to define
a new formalism neither to  develop dedicated solvers for uncertainty.
This  brings new  perspective to  tackle uncertainty  in combinatorial
optimization problems that where considered so far to be intractable.

To  show our  result is  usable  in practice,  we propose  a proof  of
concept  of our  result by  modeling a  decision-making problem  under
uncertainty in  the \microrts game  via the classical  \cop formalism,
and  we solve  it  using a  regular  \cop solver.   We  thus tackle  a
production  unit  problem  and   implement  a  bot  playing  partially
observable \microrts games and deciding what units to produce in order
to maximize its chance to counter  its opponent strategy.  Our bot has
won  the  partially   observable  track  of  the   2018  \microrts  AI
competition and  outperforms equivalent bots based  on Expected utility
or randomly producing units.

Our result  only concern  short-horizon decision-making  problems.  We
could adapt it to take into account larger horizons of action planning
and integrate it  into a bot taking long-term  strategy decision under
uncertainty.  We  also  would   like  to  investigate  problems  where
constraints  contain  stochastic  variables.   Finally,  it  would  be
interesting  to  implement  our  result  into a  bot  playing  a  more
ambitious game such as StarCraft.

\section*{Acknowledgment}

This research was supported by the Pays de la Loire region through the
Atlanstic 2020 research grant COPUL.

\bibliographystyle{abbrv}

\end{document}